\documentclass[10pt,journal]{IEEEtran}
\usepackage{amssymb,amsthm,amsmath,graphicx,cite,float,multirow,array,bm}
\usepackage[ruled]{algorithm2e}
\usepackage{threeparttable}
\usepackage{color}
\usepackage{epstopdf}
\usepackage{graphicx}

\usepackage{enumitem}
\usepackage{subfigure}
\usepackage{multirow}
\usepackage{makecell}
\usepackage{booktabs}
\usepackage[OT1]{fontenc}
\usepackage{fancyhdr}

\makeatletter

\newcommand{\Rmnum}[1]{\expandafter\@slowromancap\romannumeral #1@}

\makeatother

\begin{document}
\title{Guided Colorization Using Mono-Color Image Pairs}
\author{Ze-Hua Sheng, Hui-Liang Shen, Bo-Wen Yao, and Huaqi Zhang

\thanks{This work was supported by the National Natural Science Foundation of China under Grant 61371160. \emph{(Corresponding author: Hui-Liang Shen.)}}
\IEEEcompsocitemizethanks{\IEEEcompsocthanksitem Ze-Hua Sheng, Hui-Liang Shen, and Bo-Wen Yao are with the College of Information Science and Electronic Engineering, Zhejiang University, Hangzhou 310027, China (e-mail: shengzehua@zju.edu.cn; shenhl@zju.edu.cn; bwyao36@zju.edu.cn).}
\IEEEcompsocitemizethanks{\IEEEcompsocthanksitem Huaqi Zhang is with the vivo Mobile Communication Co., Ltd., Hangzhou 310030, China (email: zhanghuaqi@vivo.com).}
}

\IEEEcompsoctitleabstractindextext{
\begin{abstract}
Compared to color images captured by conventional RGB cameras, monochrome images usually have better signal-to-noise ratio (SNR) and richer textures due to its higher quantum efficiency. It is thus natural to apply a mono-color dual-camera system to restore color images with higher visual quality. In this paper, we propose a mono-color image enhancement algorithm that colorizes the monochrome image with the color one. Based on the assumption that adjacent structures with similar luminance values are likely to have similar colors, we first perform dense scribbling to assign colors to the monochrome pixels through block matching. Two types of outliers, including occlusion and color ambiguity, are detected and removed from the initial scribbles. We also introduce a sampling strategy to accelerate the scribbling process. Then, the dense scribbles are propagated to the entire image. To alleviate incorrect color propagation in the regions that have no color hints at all, we generate extra color seeds based on the existed scribbles to guide the propagation process. Experimental results show that, our algorithm can efficiently restore color images with higher SNR and richer details from the mono-color image pairs, and achieves good performance in solving the color bleeding problem.
\end{abstract}

\begin{IEEEkeywords}
Colorization, dual-camera system, dense scribbling, patch sampling, valid match, color ambiguity, image restoration.
\end{IEEEkeywords}
}

\maketitle

\thispagestyle{fancy}
\fancyhead{}
\lhead{This work has been submitted to the IEEE for possible publication. Copyright may be transferred without notice, after which this version may no longer be accessible.}
\chead{}
\rhead{\thepage}
\lfoot{}
\cfoot{}
\rfoot{}
\renewcommand{\headrulewidth}{0pt}
\renewcommand{\footrulewidth}{0pt}

\pagestyle{headings}

\IEEEdisplaynotcompsoctitleabstractindextext
\IEEEpeerreviewmaketitle

\section{Introduction}\label{sec:introduction}
\IEEEPARstart{M}{odern} digital cameras use a specially arranged color filter array (CFA) to capture color images on a square grid of sensors. Due to the filtering structure, each sensor can only use part of the incoming light to record color information, resulting in low signal-to-noise ratio (SNR) especially in low-light conditions. Spatial denoising techniques such as image filtering\cite{dabov2007image,elad2002origin}, sparse representation\cite{elad2006image,dong2012nonlocally}, and low-rank approximation\cite{gu2014weighted,guo2015efficient} can be applied to address this problem. However, they tend to over-smooth the details during noise removal. Recently, studies on joint filtering\cite{shibata2017misalignment,he2012guided} and image fusion\cite{nayar2000high,li2013image,connah2014spectral} indicate that, a degraded image can be better restored under the guidance of its related images of higher quality. The aforementioned adverse effect can be effectively alleviated.

Compared to RGB cameras, monochrome cameras do not have CFA structures. Hence, they cannot record color information, but can achieve higher quantum efficiency. In other words, images captured by the monochrome cameras usually contain more details and have higher SNR. It's natural that we can take advantage of a pair of related monochrome and color image to generate a color image with lower noise intensity and richer textures.

A dual-camera system can be applied to capture the mono-color image pairs. Since two cameras have different viewing angles, the image pairs have disparity. Conventional image fusion algorithms require images to be well-aligned. Therefore, previous approaches\cite{jeon2000stereo,jung2017enhancement} basically follow the scheme that they register the image pairs using stereo matching techniques at first, and then conduct the fusion process. However, when the contents of the image pair are not consistent in terms of noise and details, it's hard to yield an accurate disparity estimate, leading to several artifacts such as color bleeding.

Generally, there are two options to fuse a mono-color image pair. One option is to denoise the color image under the guidance of the monochrome image, while the other one is to colorize the monochrome image with the color one. Due to the inevitable edge smoothing of image denoising, the colorization option can obtain results with richer details. Current methods on image colorization can be categorized into scribble-based\cite{levin2004colorization,huang2005adaptive,yatziv2006fast,wang2012colorization,zhang2017real} and example-based\cite{welsh2002transferring,ironi2005colorization,gupta2012image,he2018deep,he2019progressive} ones. Scribble-based methods require users to manually add color hints to the monochrome image, then propagate those hints to the entire image. Example-based methods transfer colors from guidance to the target image, often achieved by feature matching.

In this work, we propose a novel colorization algorithm using mono-color image pairs. Considering the strong correlation between the two images, color scribbles can be created automatically through dense matching. Since our major purpose is colorization, it's actually unnecessary to align the image pair. Motivated by the commonly used assumption that adjacent pixels with similar luminance values are likely to have similar colors\cite{levin2004colorization}, we can further assume that adjacent similar structures can have similar color distributions. Hence, to colorize a pixel, we can use patches that contain this pixel and perform block matching to find their similar patches around their adjacent locations in the guidance color image. Based on the aforementioned assumption, we can directly assign the color of the matching result to the target pixel. Compared to conventional stereo matching techniques, block matching here doesn't have to check the left-right consistency, and has higher robustness when the guidance image is degraded.

Two types of outliers may occur during the scribbling process. One is the color-ambiguous pixels. Since luminance and chrominance are not always in one-to-one correspondence, even around adjacent areas, two matched pixels can be similar in luminance but totally different in color, which we define as color ambiguity. Hence, it's essential to assign each pixel multiple color candidates using overlapped patches to judge whether there is color ambiguity. The other one is the occluded pixels caused by the viewing angle difference between two images. In this work, we introduce a new concept of valid match for our dense scribbling process. That is, occlusion is directly detected based on the analysis of the color candidates, instead of using conventional methods such as left-right consistency check and ordering constraint. After removing the outliers, the rest color assignments form the initial dense scribbles.

Since we assign multiple color candidates to each pixel, an important issue is to determine how many color candidates are sufficient to produce accurate color estimates. To address this problem, we introduce a patch sampling strategy that can strike a good balance between accuracy and efficiency.

Finally, we propagate the initial scribbles to the entire image. But before propagation, we have to ensure that each region has color hints to avoid color bleeding. For the regions that have no color hints at all after dense scribbling, we generate extra color seeds based on their neighboring information.

In summary, we introduce a guided colorization algorithm to restore color images of better visual quality from the mono-color image pairs without image alignment. The main contributions of this work are as follows:
\begin{itemize}
    \item We introduce a patch-wise dense scribbling strategy with probabilistic concept based patch sampling to efficiently complete the robust color estimation.
    \item We conduct analysis on the reliability of color estimation and discuss the two outliers of the scribbling process including the occluded and the color-ambiguous pixels.
    \item We generate extra color seeds in the sparsely scribbled regions to alleviate color bleeding during color propagation.
    \item Our algorithm outperforms the state-of-the-art colorization and guided restoration algorithms to a large margin on synthetic and realistic mono-color image pairs.
\end{itemize}

\section{Related Work}\label{sec:related_work}
In this section, we briefly review the current research on image colorization and guided image enhancement.

\subsection{Image Colorization}\label{sec:image_colorization}
Generally, image colorization methods can be categorized into scribble-based and example-based ones.

\textbf{Scribble-based methods}\cite{levin2004colorization,huang2005adaptive,yatziv2006fast,wang2012colorization,zhang2017real} require users to manually add the color hints. By assuming that adjacent pixels with similar luminance values tend to have similar colors, the sparse color hints can be propagated to the entire image. This type of method was first popularized by\cite{levin2004colorization}, where colorization is modeled as an optimization problem constructed by a system of linear equations. To reduce color bleeding, \cite{huang2005adaptive} uses edge detection to prevent color hints from crossing the object boundaries, while \cite{yatziv2006fast} takes geodesic distance into consideration during color propagation. Based on the low-rank property of natural images, \cite{wang2012colorization} performs colorization by solving a matrix completion problem. In \cite{zhang2017real}, Zhang \emph{et al}. fuse color hints with semantic information through a deep neural network, which improves the colorization quality and also enables its real-time use.

\textbf{Example-based methods}\cite{welsh2002transferring,ironi2005colorization,gupta2012image,he2018deep,he2019progressive} use a set of user-provided reference images to colorize the grayscale image automatically. The most common way to make the color transfer is to establish feature correspondence between the reference image and the target grayscale image. The features can be pixels\cite{welsh2002transferring,ironi2005colorization,jeon2000stereo,he2018deep}, super-pixels\cite{gupta2012image}, or even semantic contents\cite{he2019progressive}. The only customization required is to choose an appropriate reference image. To further reduce manual efforts, \cite{he2018deep} introduces a deep neural network that can select the best reference image from a dataset and accomplish colorizing at the same time. Recently, extensive studies on the deep learning theory inspire\cite{cheng2015deep,su2020instance} that colorization can be fully automated even without reference images based on the various image priors learned by neural networks.

\begin{figure*}[tb]
  \includegraphics[scale=0.233]{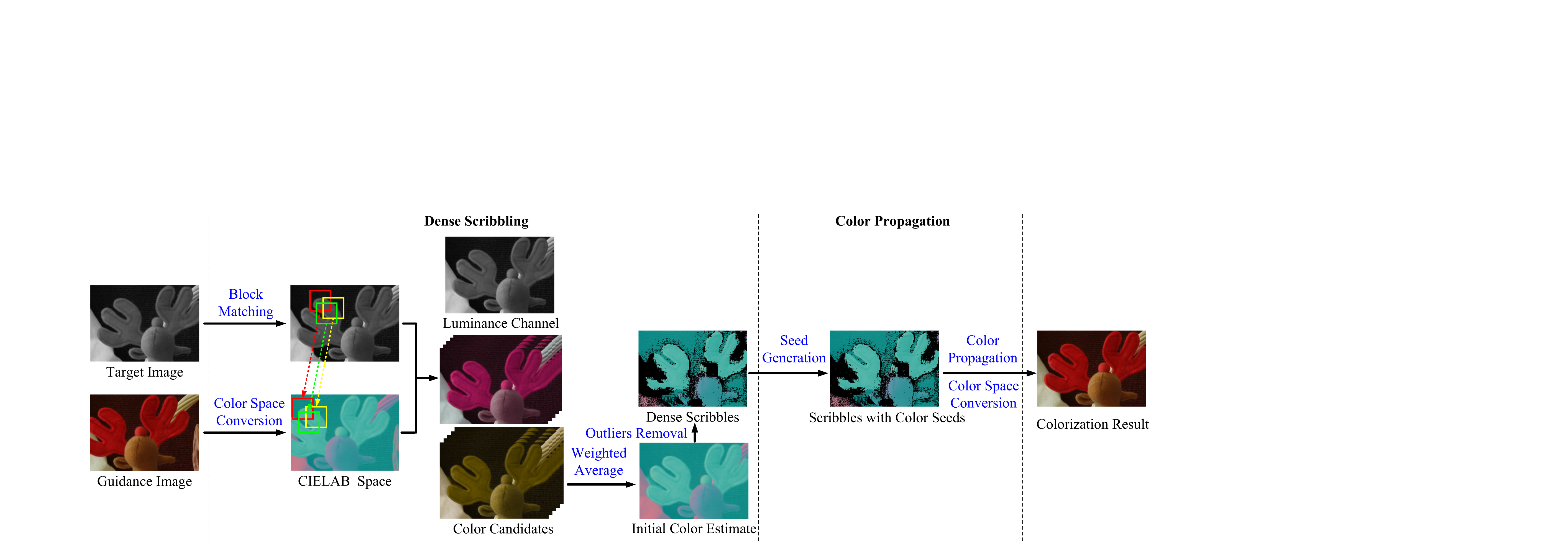}
  \caption{Framework of our colorization algorithm. First, block matching is performed between the image pair to assign color candidates. Colors are computed by weighted averaging over the candidates. Then, outliers are removed to generate the initial dense scribbles. Finally, sparse color seeds are generated in the regions that have no scribbles at all to help color hints propagate to the entire image.}
  \label{fig:framework}
\end{figure*}

\subsection{Guided Image Enhancement}\label{sec:guided_enhancement}
According to the reconstruction target, guided image enhancement can be categorized into fusion methods and restoration methods. 

\textbf{Fusion methods}\cite{li2013image,liu2016image,shen2011generalized,liu2017multi,ma2016infrared,li2018densefuse,ma2019fusiongan,connah2014spectral,wei2015hyperspectral,yang2008multimodality,bhatnagar2013directive,wang2010multi,cao2014multi,deng2021deep} aim to integrate the redundant and complementary contents from different sources, and generate enhanced images for better visual display or practical usage. The source images can be captured with different sensors, such as RGB cameras\cite{shen2011generalized,li2013image,liu2017multi}, infrared cameras\cite{ma2016infrared,li2018densefuse,ma2019fusiongan}, multi-spectral cameras\cite{connah2014spectral,wei2015hyperspectral}, or even medical cameras\cite{yang2008multimodality,bhatnagar2013directive}. They can also be obtained using the same device but with different parameter settings, \emph{e.g.}, multi-focus images\cite{wang2010multi,cao2014multi,liu2017multi}, multi-exposure images\cite{shen2011generalized,deng2021deep}, \emph{etc}. Generally, traditional fusion algorithms are conducted based on the hand-crafted features or in the transform domain\cite{li2013image,cao2014multi}. Recent deep learning techniques further improve the visual quality\cite{liu2017multi,li2018densefuse,ma2019fusiongan,deng2021deep}.

\textbf{Restoration methods}\cite{he2012guided,li2014weighted,kou2015gradient,ham2015robust,ham2017robust,shen2015mutual,guo2018mutually,wu2018fast,pan2019spatially,shibata2017misalignment,jung2017enhancement} aim to reconstruct images of higher quality from their degraded observations. In \cite{he2012guided}, He \emph{et al}. introduce the guided filter, which is the foundation of many current guided restoration methods. It leverages a local linear model to reconstruct the target image from the guidance, and can be treated as either an edge-preserving smoothing operator or a detail transfer operator. To alleviate halo artifacts and produce better edge reconstruction, \cite{li2014weighted} modifies the original cost function with an edge-aware weighting, while \cite{kou2015gradient} performs the guided filtering in the gradient domain. In some specific application scenarios such as RGB image guided depth image super-resolution, textures are not supposed to be transferred. Hence, a new type of guided filtering, called as dynamic guided filtering, is introduced to solve the problem\cite{ham2015robust,ham2017robust,shen2015mutual,guo2018mutually}. As opposed to the aforementioned static filtering process, dynamic one conducts filtering iteratively. In each iteration, the guidance image is updated to meet its structural consistency with the target image. Currently, deep learning is also applied to the guided restoration tasks. \cite{wu2018fast} trains an end-to-end neural network for fast guided filtering, while \cite{pan2019spatially} constructs a spatially variant linear representation model with learnable coefficients.

However, most of the guided restoration algorithms require images to be well-aligned. To handle misaligned image pairs, \cite{shibata2017misalignment} generates a set of translated guidances, and then performs a weighted average of the filtering results using these different guidance images. Similar to our dual-camera system settings, \cite{jung2017enhancement} aligns the mono-color pairs before filtering the noisy RGB image. The misaligned problems can be resolved to a certain extent, but there is still a lack of more in-depth analysis.

\section{Proposed Algorithm}\label{sec:proposed_algorithm}
Our algorithm contains two stages: dense scribbling and color propagation. In the dense scribbling stage, we assign monochrome pixels multiple color candidates via block matching and compute their color values by weighted averaging over the candidates. Then, we detect and remove the outliers to form the initial dense scribbles. We also introduce a patch sampling strategy to accelerate the scribbling process. In the propagation stage, we generate color seeds around the regions that have no scribbles at all, and then propagate the color hints to the entire image. The framework of our algorithm is displayed in Fig.~\ref{fig:framework}.

\subsection{Dense Scribbling}\label{sec:block_matching}
Let $\mathbf{M} \in \mathbb{R}^{H\times W}$ and $\mathbf{C} \in \mathbb{R}^{H\times W \times 3}$ denote the monochrome and the color image captured by a dual-camera system. Their maximum disparity is $D$. Our goal is to assign colors to every pixel in $\mathbf{M}$. Due to the strong correlation between the image pair, it's natural to assume that, two patches from $\mathbf{M}$ and $\mathbf{C}$ respectively that are similar in luminance are likely to have similar colors, especially within the range restricted by the maximum disparity and the epipolar constraint. Hence, $\mathbf{M}$ can be colorized automatically by searching similar patches in $\mathbf{C}$.

To begin with, we convert $\mathbf{C}$ into the CIELAB color space. The lightness channel is denoted as $\mathbf{C}_l$, describing the luminance information. Two chrominance channels are denoted as $\mathbf{C}_a$ and $\mathbf{C}_b$. As computations in $\mathbf{C}_a$ and $\mathbf{C}_b$ are identical, we refer to $\mathbf{C}_a$ as the chrominance channel below for brevity.

We divide $\mathbf{M}$ into overlapped patches of size $S \times S$. For a monochrome patch $\mathbf{p} \in \mathbb{R}^{S\times S}$, we search for its most similar color patch in $\mathbf{C}$, within a window of size $W_h \times W_v$. In this work, we set $S=16$, the horizontal search range $W_h=D$, and the vertical search range $W_v=30$. The difference between $\mathbf{p}$ and a color patch $\mathbf{q}\in\mathbb{R}^{S\times S\times 3}$ is measured by their Euclidean distance in luminance, which is computed as
\begin{equation}\label{eq:l2_distance}
D_l(\mathbf{p}, \mathbf{q}) = ||\mathbf{p} - \mathbf{q}_l||_F^2,
\end{equation}
where $\mathbf{q}_l$ is the lightness channel of $\mathbf{q}$.

Then, we select $\mathbf{q}$ with the smallest distance, and combine its chrominance channel $\mathbf{q}_a$ with $\mathbf{p}$ to assign each monochrome pixel a color candidate. Since all the patches are overlapped, a pixel can be included in different patches, and can thus obtain multiple color candidates through block matching of different patches.

We stack all the color candidates into a 3D tensor $\mathbf{U} \in \mathbb{R}^{H\times W \times K}$, where $K$ is the number of candidates of a pixel and $K_\mathrm{max}=S^2$. Luminance values of the matched pixels are stacked in $\mathbf{L}\in\mathbb{R}^{H\times W \times K}$. A weighted average is then performed across these candidates to obtain the color estimates. The chrominance value of $\mathbf{M}$ is thus computed by
\begin{equation}\label{eq:weighted_average}
\mathbf{M}_a(i,j) = \sum_{k=1}^K \mathbf{W}(i,j,k) \cdot \mathbf{U}(i,j,k),
\end{equation}
where $\mathbf{W}\in\mathbb{R}^{H\times W \times K}$ is the weight tensor.

Based on the assumption that adjacent pixels with similar luminance values are similar in color, we determine the value of each element in $\mathbf{W}$ by computing the absolute luminance difference between the target monochrome pixel and the matched pixel, computed as
\begin{equation}\label{eq:weight}
\mathbf{W}(i,j,k) = \mathbf{Z}(i,j) \cdot \frac{1}{|\mathbf{L}(i,j,k) - \mathbf{M}(i,j)| + \varepsilon},
\end{equation}
where $\mathbf{Z}$ is the normalization matrix that ensures $\sum_{k=1}^K \mathbf{W}(i,j,k) = 1$, and $\varepsilon$ is set to a small value to prevent the denominator from being zero.

In addition, random noise in the guidance image can affect the accuracy of dense scribbling, and can also cause noise residue in the colorization results. A simple and fast pre-denoising on the guidance image can help to alleviate the aforementioned problems directly. In this work, we apply the random redundant DCT (RRDCT) denoising algorithm introduced in~\cite{fujita2015randomized} for noise suppression, which runs almost in real time. Although our algorithm already has some robustness to random noise due to the patch-wise matching scheme, the pre-denoising step can help to further improve the colorization quality, especially when the noise level is relatively high.

\begin{figure}[t]
\centering
\includegraphics[scale=0.79]{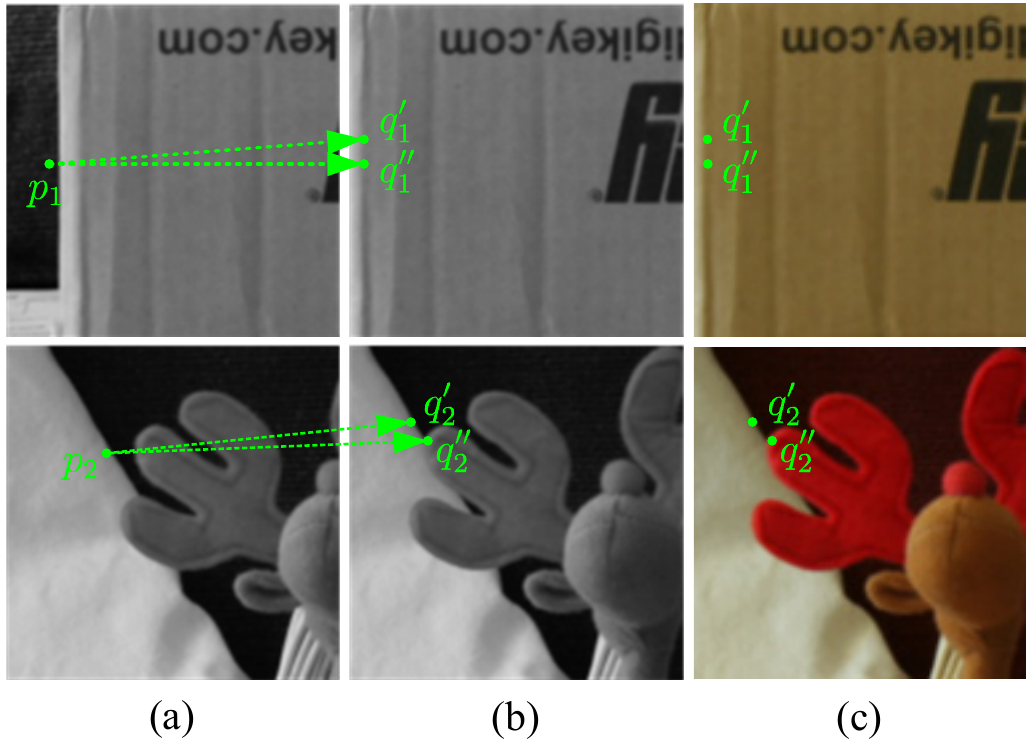}
\caption{Examples of two types of outliers. (a) Monochrome target image. (b) Lightness channel of the guidance image. (c) Guidance image. $p_1$ is the occluded pixel, whose luminance value is $0.11$. But the luminance values of its matched pixels $q'_1$ and $q''_1$ are $0.45$ and $0.46$, respectively. $p_2$ is the color-ambiguous pixel. The luminance values of $p_2$ and its matched pixels $q'_2$ and $q''_2$ are all $0.44$, but $q'_2$ and $q''_2$ have different colors.}
\label{fig:outliers}
\end{figure}

Color assignment via block matching still remains two problems to be solved. First, the algorithm can be expensive in both time and space if all the patches are used in block matching, which proves to be unnecessary in Section~\ref{sec:patch_sampling}. Second, patches from the occluded regions may not be able to find their similar matches, resulting in wrong color assignment. In addition, since luminance and chrominance are not always in one-to-one correspondence, color candidates of one pixel may differ from each other, which we describe as color ambiguity. Fig.~\ref{fig:outliers} shows examples of two outliers respectively. Therefore, colors of those outliers should be removed and re-estimated. Then, the remaining colors form the initial dense scribbles. Section~\ref{sec:patch_sampling} and ~\ref{sec:outliers_detection} will focus on how to address the two problems respectively.

\subsection{Patch Sampling}\label{sec:patch_sampling}
During block matching, we use patches of size $S\times S$ to allocate $S^2$ color candidates to each pixel. However, it can be time-consuming when the size of image is quite large. A proper sampling strategy is essential to accelerate the scribble-generating process.

From the perspective of a pixel, patch sampling is equivalent to extracting $N$ colors from the overall $S^2$ candidates, under a sampling rate of $\rho=N/S^2$. If $T$ of them already have similar luminance values to the target pixel, we can assert that it's enough to give an accurate color estimate and can terminate the matching process. To strike a good balance between efficiency and accuracy, the sampling-related values of $N$ and $T$ should be carefully selected, where $N$ determines the sampling rate and $T$ indicates whether the color candidates are sufficient to generate a correct color estimate.

For clarity, the main notations and their corresponding meanings used in patch sampling are listed in Table~\ref{tab:notations}. Suppose that we extract $N$ values from the $S^2$ color candidates of pixel $p$. We denote it as event $\mathcal{A}_r$ if $r$ of them have similar luminance values to $p$, measured by their absolute differences. A threshold is required to judge whether two pixels are similar or not. If their absolute luminance difference is lower than the threshold, we can assert that they are similar. In this work, we use the just-noticeable difference (JND) threshold introduced in~\cite{wu2016enhanced}, which is a  perceptual threshold computed as
\begin{equation}\label{eq:weight}
\mathrm{J}(p) = J_L(p) + J_T(p) - 0.3 \min \left(J_L(p), J_T(p)\right),
\end{equation}
where $J_L(p)$ is the intensity term and $J_T(p)$ is the texture term.

\begin{figure}[t]
\includegraphics[scale=0.3]{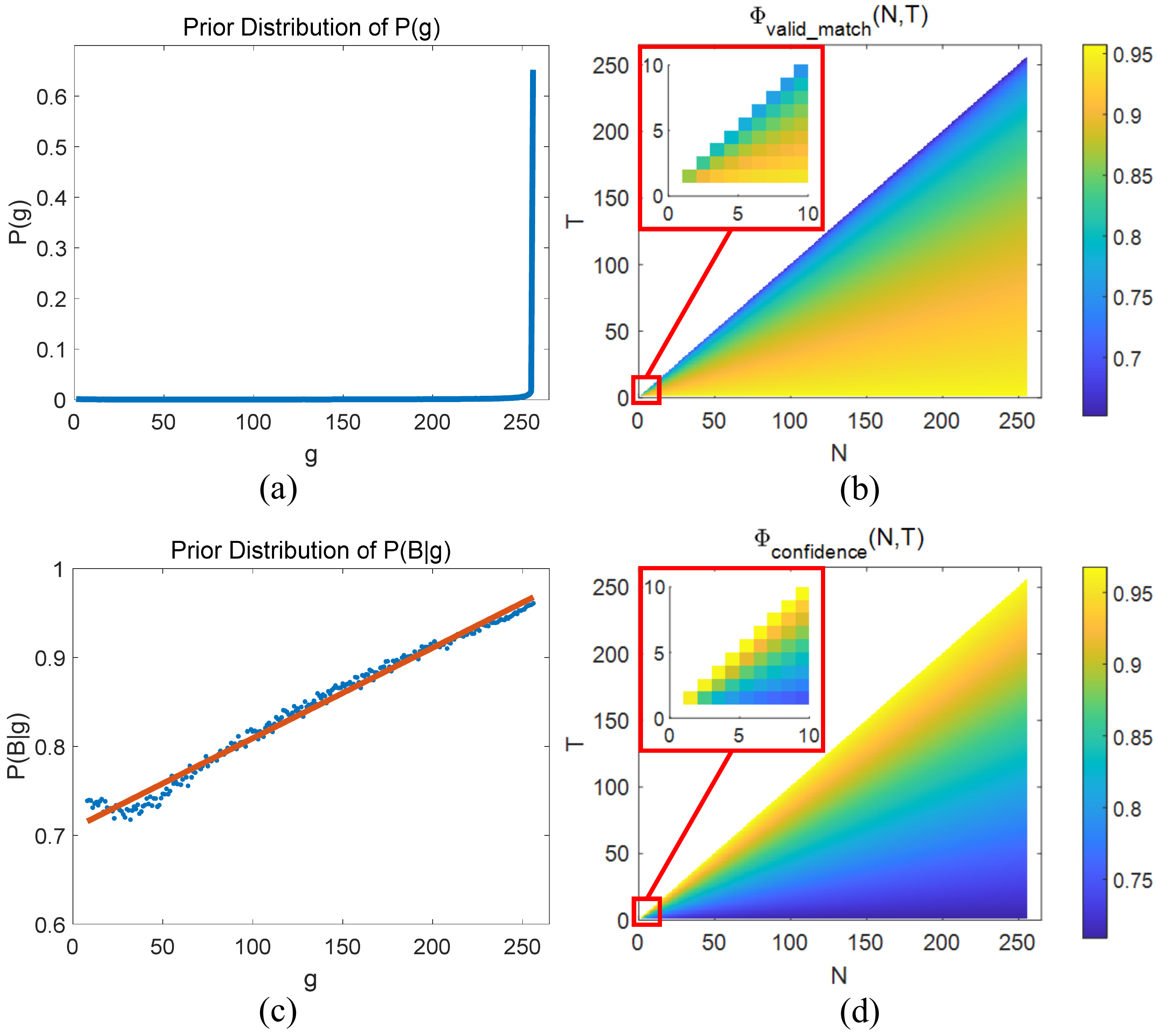}
\caption{Visualization of two criteria $\Phi_\mathrm{valid\_match}(N,T)$ and $\Phi_\mathrm{confidence}(N,T)$ for patch sampling. (a) and (c) display the prior distribution of $P(g)$ and $P(\mathcal{B}|g)$ respectively. (b) and (d) display the values of $\Phi_\mathrm{valid\_match}(N,T)$ and $\Phi_\mathrm{confidence}(N,T)$ under different settings of $N$ and $T$.}
\label{fig:priors}
\end{figure}

Denoting $g$ as the total number of candidates that are similar to $p$ judged by JND without patch sampling, its prior probability distribution $P(g), \ 1\leq g \leq S^2$, is depicted in Fig.~\ref{fig:priors} (a), which is a statistical result obtained by massive experiments on the Middlebury datasets~\cite{scharstein2007learning,hirschmuller2007evaluation}. Since most of the image contents are flat, majority of the pixels are likely to obtain similar color candidates. Color information obtained by block matching is quite redundant, proving that using all the patches to generate an accurate color estimate is a thankless job.

\begin{table}[tb]
\renewcommand\arraystretch{1.3}
\renewcommand\tabcolsep{7pt}
\centering
\caption{Main notations used in patch sampling.}\label{tab:notations}
\begin{tabular}{m{40 pt}<{\centering}|m{170 pt}}
  \hline\hline
  Notations&\makecell[c]{Meanings} \\\hline
  $g$ & Number of similar matched pixels without patch sampling \\\hline
  $r$ & Number of similar matched pixels with patch sampling \\\hline
  $\mathcal{A}_r,\mathcal{A}_T$ & Event of matching $r,T$ similar pixels with patch sampling \\\hline
  $\mathcal{B}$ & Event of obtaining correct color estimate \\\hline
  $P(g)$ & Prior probability of matching $g$ similar pixels without patch sampling \\\hline
  $P(\mathcal{A}_r)$ & Probability of matching $r$ similar pixels with patch sampling  \\\hline
  $P(\mathcal{B}|g)$ & Prior probability of correct color estimation when matching $g$ similar pixels without patch sampling \\\hline
  $P(\mathcal{B}|\mathcal{A}_T)$ & Posteriori probability of correct color estimation when matching $T$ similar pixels with patch sampling \\\hline
  \hline
\end{tabular}
\end{table}

We define that $p$ achieves a valid match if $r\geq T$. Valid matches do not require extra matching for color estimation. A good sampling strategy should guarantee a high probability of valid matches, otherwise the scribbles may become sparse, which is not conducive to accurate colorization. The probability of valid match is computed by
\begin{equation}\label{eq:prob_valid_match}
\Phi_\mathrm{valid\_match}(N,T) = \sum_{r=T}^N P(A_r),
\end{equation}
where
\begin{equation}\label{eq:prob_A_R}
\begin{aligned}
P(A_r) &= \sum_{g=1}^{S^2} P(A_r, g) = \sum_{g=1}^{S^2} P(g) \cdot P(A_r|g) \\
&= \sum_{g=1}^{S^2} P(g) \cdot \frac{\mathrm{C}^r_g\cdot \mathrm{C}^{N-r}_{S^2-g}}{\mathrm{C}^N_{S^2}},
\end{aligned}
\end{equation}
and $\mathrm{C}_m^n$ denotes the combination function.

The second aspect to judge a good sampling strategy is the confidence that a valid match generates a correct color estimate. Due to color ambiguity, the extracted candidates may be correct in luminance but wrong in color. We define event $\mathcal{B}$ to describe that the color scribble computed by the weighted average of the color candidates is correct, also judged by JND. The prior distribution of scribbling accuracy $P(\mathcal{B}|g)$ is shown in Fig.~\ref{fig:priors} (c), which is also a statistical result obtained via experiments on the Middlebury datasets. The result shows that color ambiguity occurs more often where color candidates are less consistent in luminance. For computation convenience, the prior distribution $P(\mathcal{B}|g)$ can be approximately fitted by a linear model. Thus, the confidence of valid match $\Phi_\mathrm{confidence}(N,T)$ can be measured by the posteriori probability $P(\mathcal{B}|\mathcal{A}_T)$, computed as
\begin{equation}\label{eq:confidence}
\begin{aligned}
\Phi_\mathrm{confidence}(N,T) &= P(\mathcal{B}|\mathcal{A}_T) =  \frac{P(\mathcal{A}_T, \mathcal{B})}{P(\mathcal{A}_T)} \\
&= \frac{\sum_{g=1}^{S^2}P(\mathcal{A}_T, \mathcal{B}|g)P(g)}{P(\mathcal{A}_T)} \\
&= \frac{\sum_{g=1}^{S^2}P(\mathcal{A}_T|g)P(\mathcal{B}|g)P(g)}{P(\mathcal{A}_T)}.
\end{aligned}
\end{equation}

Values of $\Phi_\mathrm{valid\_match}(N,T)$ and $\Phi_\mathrm{confidence}(N,T)$ are visualized in Fig.~\ref{fig:priors} (b) and (d). We apply these two criteria to determine the values of $N$ and $T$. A good sampling process should achieve high probability of valid match to generate dense scribbles, and also have high confidence in valid match to compute colors correctly. According to Fig.~\ref{fig:priors}, by setting $N=4$ and $T=4$, we can get $\Phi_\mathrm{valid\_match}(N,T)=0.7920$ and $\Phi_\mathrm{confidence}(N,T)=0.9639$, which means $1/64$ of the patches are already sufficient to colorize $79.2\%$ of the pixels with confidence $0.9639$. 

After determining the sampling rate, we extract patches by setting a constant stride. Patches can also be randomly selected through sampling techniques such as Poisson disk sampling\cite{cook1986stochastic,dunbar2006spatial}, which achieves similar colorization results.

\subsection{Outliers Removal}\label{sec:outliers_detection}
The major reasons that cause wrong color estimation are occlusion and color ambiguity. To yield a colorization result of higher visual quality, it's essential to detect these outliers, remove their initial color estimates, and re-colorize them using their adjacent information.

Occluded pixels can be easily detected in the block matching stage, without conventional left-right consistency check. Using the sampling strategy mentioned above, pixels that fail to achieve valid matches can be recognized as occluded pixels.

To check whether a pixel that completes valid match has ambiguous color candidates, we sort its color values in ascend order. If there exists an absolute difference between two adjacent values that is greater than the threshold $\tau$, we can assert that the chrominance values of its color candidates are not consistent. In our work, leveraging a constant $\tau$ is enough to give plausible detection results of ambiguous-color pixels. Here, $\tau$ is set to $5/255$.

\subsection{Seed Generation and Color Propagation}
After obtaining the scribbles, we propagate them to the entire image to generate the complete colorization result. Color propagation is one of the key problems in the study of scribble-based colorization algorithms, which basically follow the assumption that adjacent pixels with similar luminance values are likely to have similar colors. Hence, the existed color hints can gradually spread to their adjacent pixels with similar luminance, until all the pixels are colorized. Unlike the sparse, manually created color hints, our scribbles are dense and are less likely to cause color bleeding during propagation. The pioneering work~\cite{levin2004colorization} introduced by Levin \emph{et al}. is already sufficient to propagate colors properly. The only concern lies in the large occluded regions where there are no color hints at all. To further alleviate color bleeding in these regions, we generate extra color seeds to guide the propagation process.

\begin{figure}[t]
\includegraphics[scale=1]{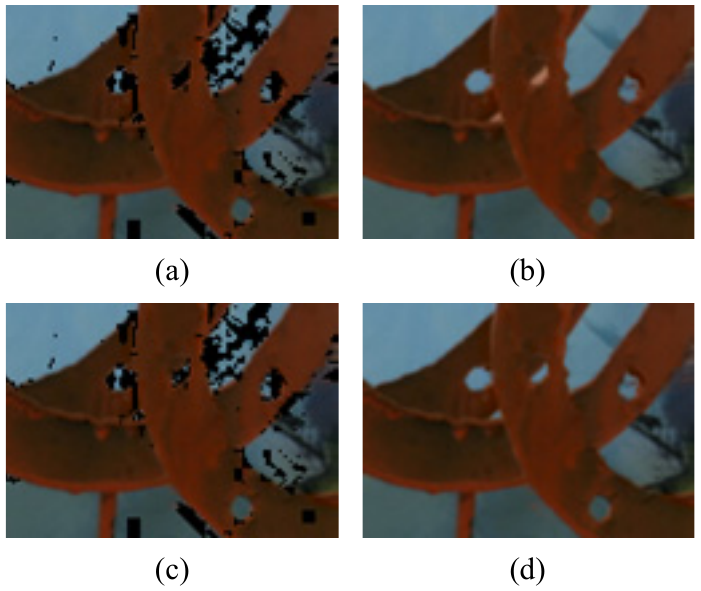}
\centering
\caption{Color propagation results. (a) Dense scribbles without color seeds. (b) Colorization result without color seeds. (c) Dense scribbles with color seeds. (d) Colorization result with color seeds.}
\label{fig:color_seeds}
\end{figure}

To begin with, the monochrome image is divided into non-overlapped blocks of size $B\times B$. In each block, we categorize pixels into several luminance levels. Specifically, we sort their luminance values in ascend order. A level splitter is set where the absolute luminance difference between two adjacent values is greater than the threshold $\tau_l$. We have to ensure that each luminance level of pixels has at least one color hint. Otherwise, we select one pixel $p$ from that level as the seed pixel, and search for $N_p$ of its most similar pixels in luminance, denoted as $q_n$, $1\leq n \leq N_p$, in the initial scribbles within a window of size $W_N\times W_N$. Here, we set $B=20$, $\tau_l=9/255$, $N_p=3$ and $W_N=50$. The color of $p$ is computed as
\begin{equation}\label{eq:seed_color}
p_a = \sum_{n=1}^{N_p} z_p \cdot \frac{1}{|p-q_{l,n}|+\varepsilon} \cdot q_{a,n},
\end{equation}
where $q_{l,n}$ and $q_{a,n}$ denote the luminance value and the chrominance value of pixel $p_n$ respectively, and $z_p$ denotes the normalization parameter. With the initial dense scribbles and the extra color seeds, we apply the color propagation algorithm~\cite{levin2004colorization} to colorize the remaining pixels. Fig.~\ref{fig:color_seeds} (b) and (d) compare the colorization results with and without color seeds, showing that color seeds help to reduce incorrect color propagation.

In practice, extra color seeds only account for very small part of the total pixels ($<0.05\%$ in the Middlebury dataset), and have little impact on the overall accuracy. To estimate more reliable colors for them, we use a window centered at each seed pixel rather than using just a single pixel for matching. Sometimes the color seeds may be incorrect in the fully occluded regions, but fetching color from neighbors can make the occluded contents locally consistent in chrominance. 

\section{Experiments}
In this section, we first introduce the experimental setup. Then, we validate our algorithm by performing it step-by-step to show the contributions of each stage. Next, comparisons with different methods are conducted on both synthetic and real-world mono-color image pairs. Finally, we have a discussion on several implementation details.

\subsection{Experimental Setup}
Our algorithm is implemented in MATLAB R2020b. All of our experiments are conducted on a personal computer with 2.70GHz CPU (Intel Core i5-6400) and 24 GB RAM.

We evaluate our algorithm on the public Middlebury 2005, 2006 datasets~\cite{scharstein2007learning,hirschmuller2007evaluation} and a SceneFlow test set~\cite{mayer2016large}. The Middlebury datasets consist of 30 groups of multi-view color images for stereo matching evaluation. In our experiments, we convert the left view image into the CIELAB color space, and take the lightness channel as the target monochrome image. Its adjacent right view is taken as the guidance image. The original left-view color image is taken as ground truth. The SceneFlow test set contains 120 stereo images, where the synthetic image pairs are generated in the same way. Two common quality metrics, \emph{i.e.}, peak signal-to-noise ratio (PSNR) and structural similarity index (SSIM)~\cite{wang2004image}, are employed to assess the colorization performance.

We also compare our algorithm to the state-of-the-art colorization algorithms and guided restoration algorithms. The colorization algorithms include Jeon16~\cite{jeon2000stereo}, He18~\cite{he2018deep}, He19~\cite{he2019progressive}, Jampani17~\cite{jampani2017video}, and Zhang19~\cite{zhang2019deep}. Jeon16~\cite{jeon2000stereo} uses stereo matching to register the mono-color image pair before colorization. He18~\cite{he2018deep} completes matching and colorization with two separated neural networks. Its following work, Zhang19~\cite{zhang2019deep}, is applied to video colorization tasks. He19~\cite{he2019progressive} is also a deep learning algorithm that transfers color from guidance to target by constructing semantic correspondence. Jampani17~\cite{jampani2017video} is a video colorization algorithm that aims to propagate colors frame by frame from the reference one. Its application scenario is similar to ours, for the mono-color pairs can be recognized as two adjacent video frames. The guided restoration algorithms include Pan19~\cite{pan2019spatially}, Shibata17~\cite{shibata2017misalignment}, and Jung17~\cite{jung2017enhancement}. Since their target images are the RGB ones, we only conduct visual comparisons.

To validate our algorithm in real scenes, we set up a dual-camera system and create our own mono-color image pairs. The monochrome camera captures the left-view image, while the color camera captures the right view. Since the image pairs do not have ground truth, we only conduct the visual assessment as well.

\subsection{Performance Assessment}
\begin{figure}[t]
\centering
\includegraphics[scale=0.415]{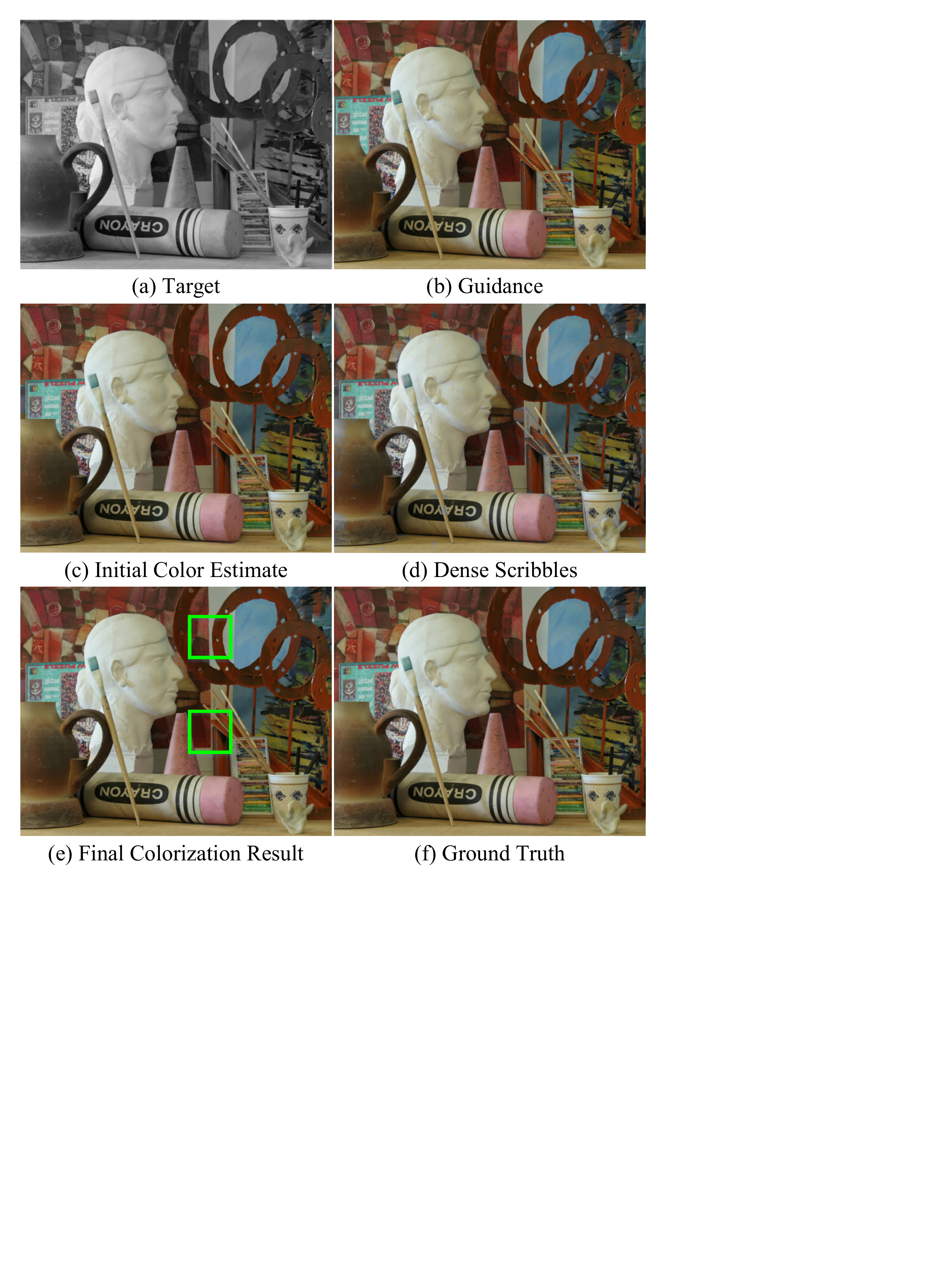}
\caption{Colorization results of each step in our algorithm. Compared with the initial estimate, wrong colors are effectively corrected in the final colorization result.}
\label{fig:performance}
\end{figure}

To validate our colorization algorithm, we perform it step by step and show the contributions of each stage to the colorization results.
In the dense scribbling stage, block matching allocates each monochrome pixel an initial color value. A shown in Fig.~\ref{fig:performance} (c), the initial colors are already plausible in most regions, even in some occluded areas. With outliers detection, the wrong colorization areas are effectively located. We remove these incorrect color assignments and obtain the dense scribbling, displayed in Fig.~\ref{fig:performance} (d). Fig.~\ref{fig:performance} (e) shows the final colorization result after color propagation.

\begin{figure}[t]
\centering
\includegraphics[scale=0.425]{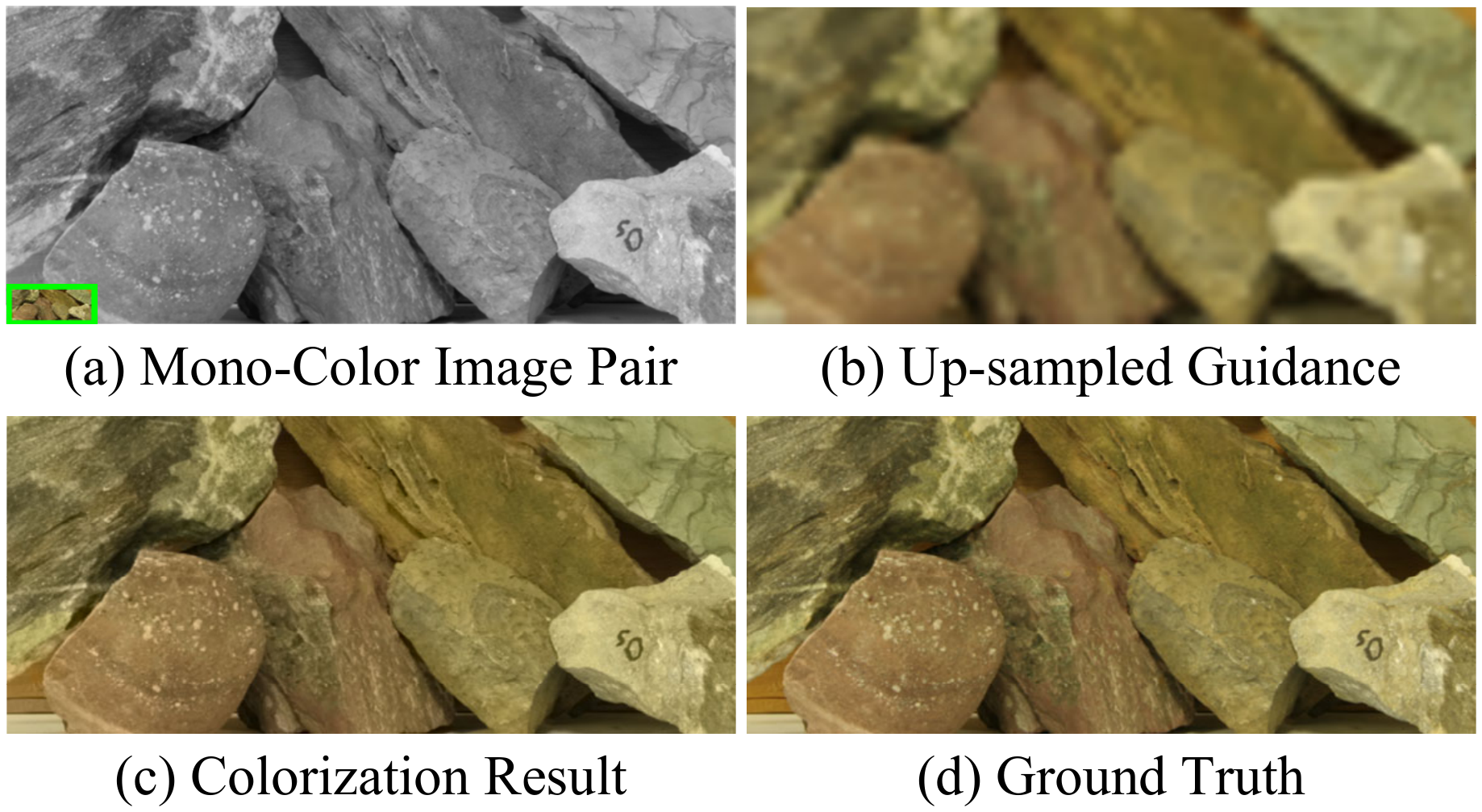}
\caption{Colorization result when the size of the target image is $64\times$ the size of the guidance image.}
\label{fig:upsampling}
\end{figure}

In the real application scenarios, the resolutions of the monochrome image and the color image can be different. Hence, we evaluate our algorithm in the situation where the color image is smaller than the monochrome image. Before colorization, we up-sample the guidance image using bicubic interpolation to match the size of the target image for block matching. As shown in Fig.~\ref{fig:upsampling} (b), the up-sampled guidance image is over-smooth, but we can still achieve good colorization results thanks to the robustness of block matching. Similarly, over-smoothing caused by image denoising on the guidance image also has little adverse effect on the colorization accuracy.

\begin{figure}[t]
\centering
\includegraphics[scale=0.33]{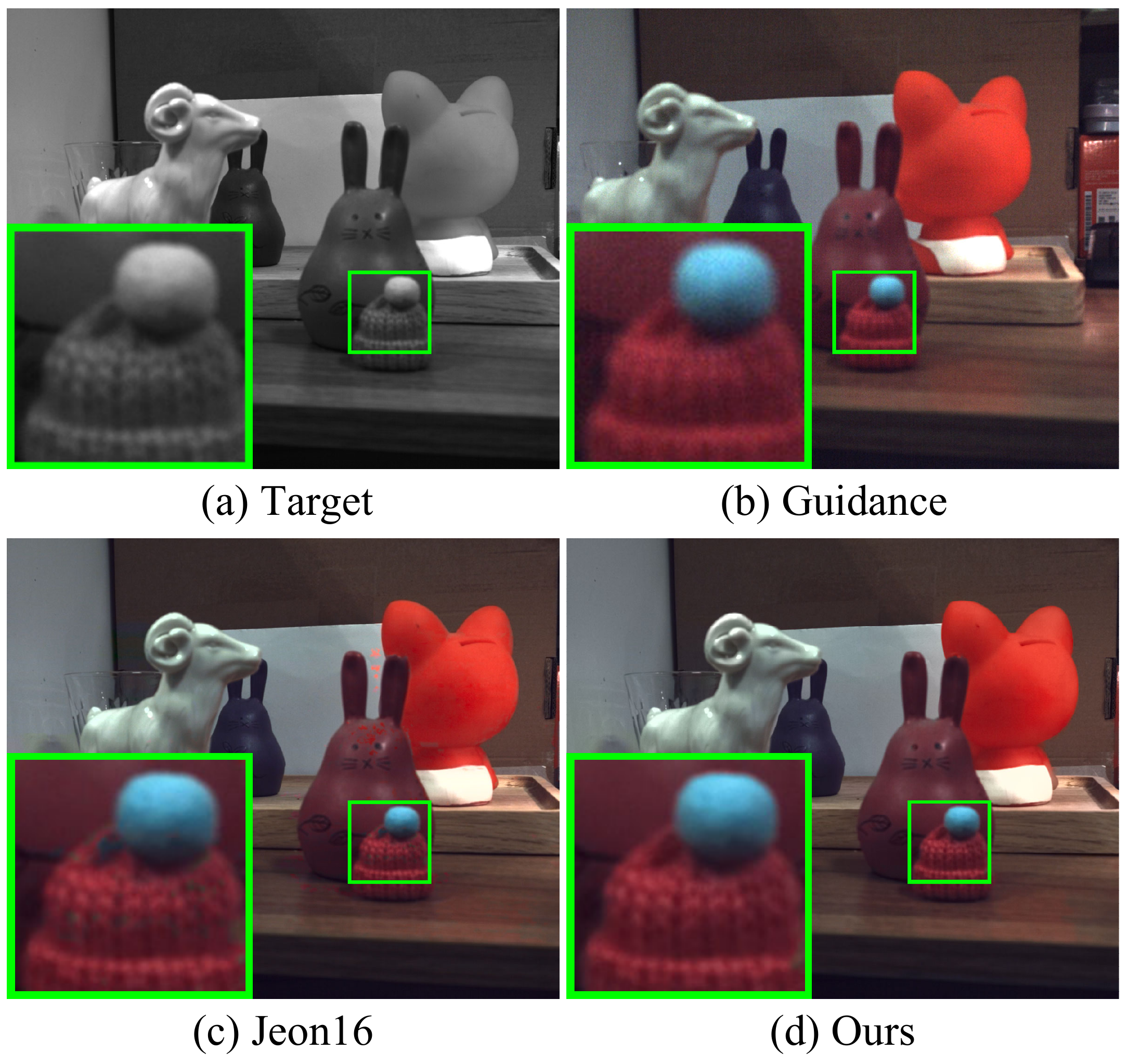}
\caption{Colorization result of a realistic mono-color image pair. The colorization results are obtained by Jeon16~\cite{jeon2000stereo} and our algorithm, respectively.}
\label{fig:real_scene}
\end{figure}

We also capture mono-color image pairs with a dual-camera system to evaluate our algorithm in real scenarios. Due to the filtering structure of RGB cameras, intensities of the color images are lower than the monochrome images. The color image requires to match the intensity level of the monochrome image for accurate feature matching. Let $\overline{\mathbf{C}}$ denote the grayscale version of the color image $\mathbf{C}$ obtained by averaging its R, G and B channels. Intensities of $\overline{\mathbf{C}}$ and the monochrome image $\mathbf{M}$ have a linear relationship, \emph{i.e.}, $\mathbf{M}$ and $\lambda \overline{\mathbf{C}}$ are of the same intensity level. The linear parameter $\lambda$ is computed as the average luminance ratio between the image pair. Hence, block matching can be performed between $\mathbf{M}$ and $\overline{\mathbf{C}}'= \lambda \overline{\mathbf{C}}$ for realistic mono-color image pairs. Here, $\lambda$ can also be computed locally for each search region to achieve higher robustness. It's worth noting that the matching process is conducted on the raw data. To handle non-linear situations in the image signal processing pipeline such as Gamma transformation, the linear model still works well.

Fig.~\ref{fig:real_scene} (d) shows one of our colorization results. Compared to the guidance image displayed in Fig.~\ref{fig:real_scene} (b), whose intensity level has been matched to the target image, our restored color image has lower noise intensities and contains richer textures that conventional denoising algorithms cannot obtain. Fig.~\ref{fig:real_scene} (c) is obtained by Jeon16~\cite{jeon2000stereo} that registers the image pair via stereo matching and then perform colorization. Its colorization result suffers from the color bleeding problem. One of the reasons is that it's hard to estimate an accurate disparity map when the contents of the image pair are not consistent in terms of noise and textures. In comparison, our algorithm is more robust and can restore color images with better visual quality in terms of less color bleeding, demonstrating that image alignment is not an essential step for accurate colorization using mono-color image pairs.

\begin{table*}[!htb]\scriptsize
\renewcommand\arraystretch{1.5}
\renewcommand\tabcolsep{7pt}
\centering
\caption{The average PSNR (dB) and SSIM values of different algorithms on image pairs from the Middlebury datasets.}\label{tab:quantitative_result}
\begin{tabular}{p{60 pt}|p{37 pt}<{\centering}|p{37 pt}<{\centering}|p{37 pt}<{\centering}|p{37 pt}<{\centering}|p{37 pt}<{\centering}|p{37 pt}<{\centering}|p{37 pt}<{\centering}|p{37 pt}<{\centering}}
  \hline\hline
  \multirow{3}{*}{} & \multicolumn{4}{c|}{$\alpha=0$} & \multicolumn{4}{c}{$\alpha=0.03^2$} \\\cline{2-9}
  &\multicolumn{2}{c|}{$\sigma^2=0$} & \multicolumn{2}{c|}{$\sigma^2=0.03^2$} & \multicolumn{2}{c|}{$\sigma^2=0$} & \multicolumn{2}{c}{$\sigma^2=0.03^2$} \\\cline{2-9}
  &PSNR&SSIM&PSNR&SSIM&PSNR&SSIM&PSNR&SSIM \\\hline
  Jeon16~\cite{jeon2000stereo} & 39.00 & 0.9922 & 37.48 & 0.9892 & 36.68 & 0.9862 & 35.98 & 0.9839\\\hline
  He18~\cite{he2018deep} & 33.36 & 0.9740 & 32.35 & 0.9670 & 33.23 & 0.9720 & 31.93 & 0.9645 \\\hline
  He19~\cite{he2019progressive} & 29.75 & 0.9414 & 28.89 & 0.9323 & 29.49 & 0.9406 & 28.76 & 0.9320 \\\hline
  Zhang19~\cite{zhang2019deep} & 29.49 & 0.9546 & 29.42 & 0.9538 & 29.50 & 0.9544 & 29.46 & 0.9539 \\\hline
  Jampani17~\cite{jampani2017video} & 23.34 & 0.7284 & 23.25 & 0.7269 & 23.28 & 0.7275 & 23.22 & 0.7261 \\\hline
  Ours & \textbf{41.50} & \textbf{0.9954} & \textbf{40.30} & \textbf{0.9938} & \textbf{38.81} & \textbf{0.9909} & \textbf{38.56} & \textbf{0.9904} \\\hline\hline

  \hline
\end{tabular}
\end{table*}

\begin{table*}[!htb]\scriptsize
  \renewcommand\arraystretch{1.5}
  \renewcommand\tabcolsep{7pt}
  \centering
  \caption{The average PSNR (dB) and SSIM values of different algorithms on image pairs from the SceneFlow test set.}\label{tab:quantitative_result_sf}
  \begin{tabular}{p{60 pt}|p{37 pt}<{\centering}|p{37 pt}<{\centering}|p{37 pt}<{\centering}|p{37 pt}<{\centering}|p{37 pt}<{\centering}|p{37 pt}<{\centering}|p{37 pt}<{\centering}|p{37 pt}<{\centering}}
    \hline\hline
    \multirow{3}{*}{} & \multicolumn{4}{c|}{$\alpha=0$} & \multicolumn{4}{c}{$\alpha=0.03^2$} \\\cline{2-9}
    &\multicolumn{2}{c|}{$\sigma^2=0$} & \multicolumn{2}{c|}{$\sigma^2=0.03^2$} & \multicolumn{2}{c|}{$\sigma^2=0$} & \multicolumn{2}{c}{$\sigma^2=0.03^2$} \\\cline{2-9}
    &PSNR&SSIM&PSNR&SSIM&PSNR&SSIM&PSNR&SSIM \\\hline
    Jeon16~\cite{jeon2000stereo} & 31.50 & 0.9578 & 30.27 & 0.9340 & 30.77 & 0.9467 & 29.99 & 0.9281\\\hline
    He18~\cite{he2018deep} & 29.16 & 0.9207 & 29.34 & 0.9243 & 29.26 & 0.9215 & 29.37 & 0.9249 \\\hline
    He19~\cite{he2019progressive} & 26.07 & 0.8400 & 25.89 & 0.8298 & 26.08 & 0.8383 & 25.87 & 0.8291 \\\hline
    Zhang19~\cite{zhang2019deep} & 27.47 & 0.9012 & 27.46 & 0.8991 & 27.47 & 0.9006 & 27.45 & 0.8982 \\\hline
    Jampani17~\cite{jampani2017video} & 25.25 & 0.7865 & 25.06 & 0.7843 & 25.05 & 0.7843 & 23.96 & 0.7487 \\\hline
    Ours & \textbf{34.30} & \textbf{0.9728} & \textbf{33.19} & \textbf{0.9598} & \textbf{33.62} & \textbf{0.9657} & \textbf{32.97} & \textbf{0.9575} \\\hline\hline
  
    \hline
  \end{tabular}
  \end{table*}

\begin{figure*}[h]
\centering
\includegraphics[scale=0.347]{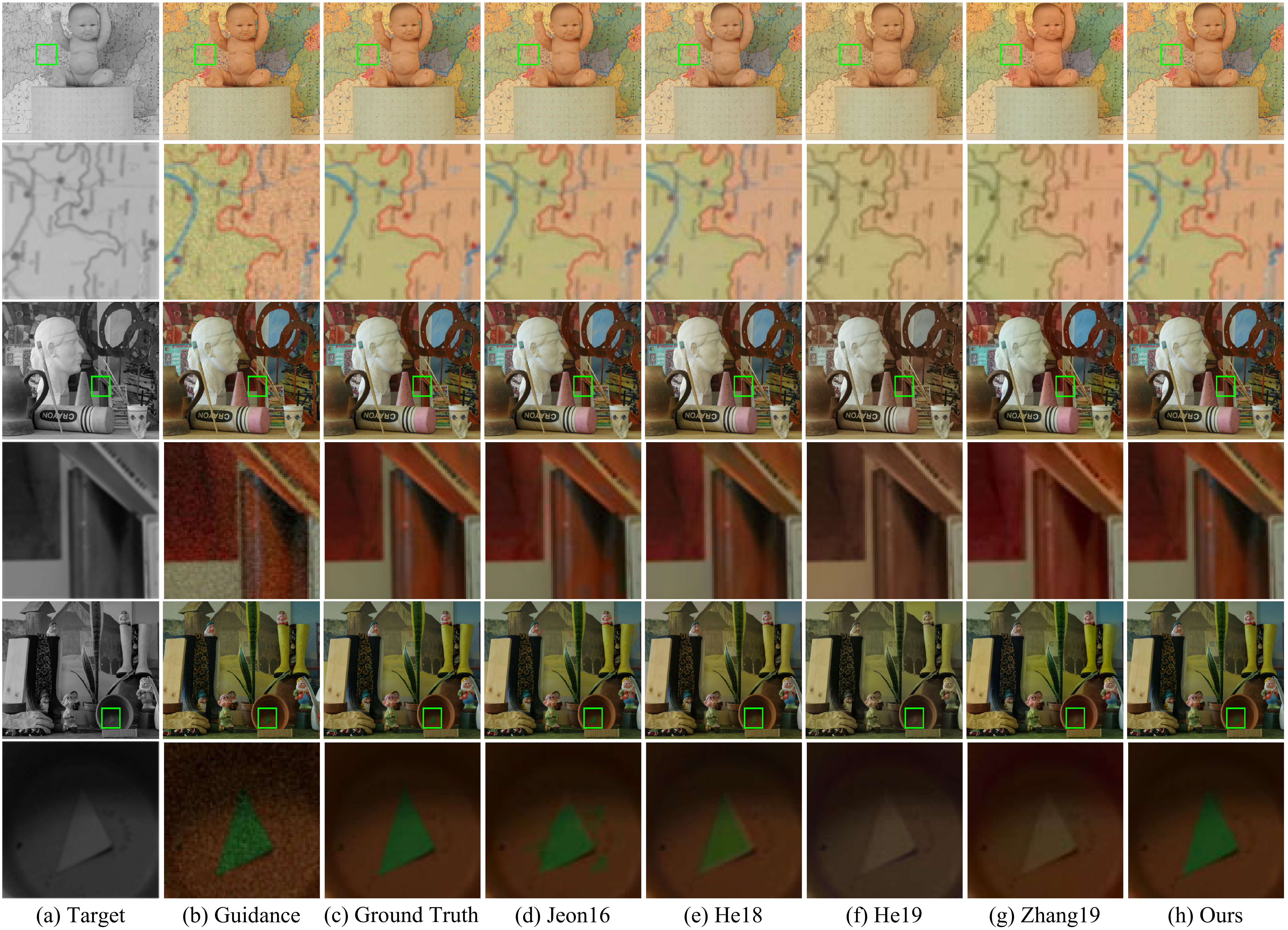}
\caption{Visual comparison of colorization results on the image pairs from the Middlebury datasets, obtained by Jeon16~\cite{jeon2000stereo}, He18~\cite{he2018deep}, He19~\cite{he2019progressive}, Zhang19~\cite{zhang2019deep} and our algorithm, respectively.}
\label{fig:comp}
\end{figure*}

\begin{figure}[h]
\centering
\includegraphics[scale=0.454]{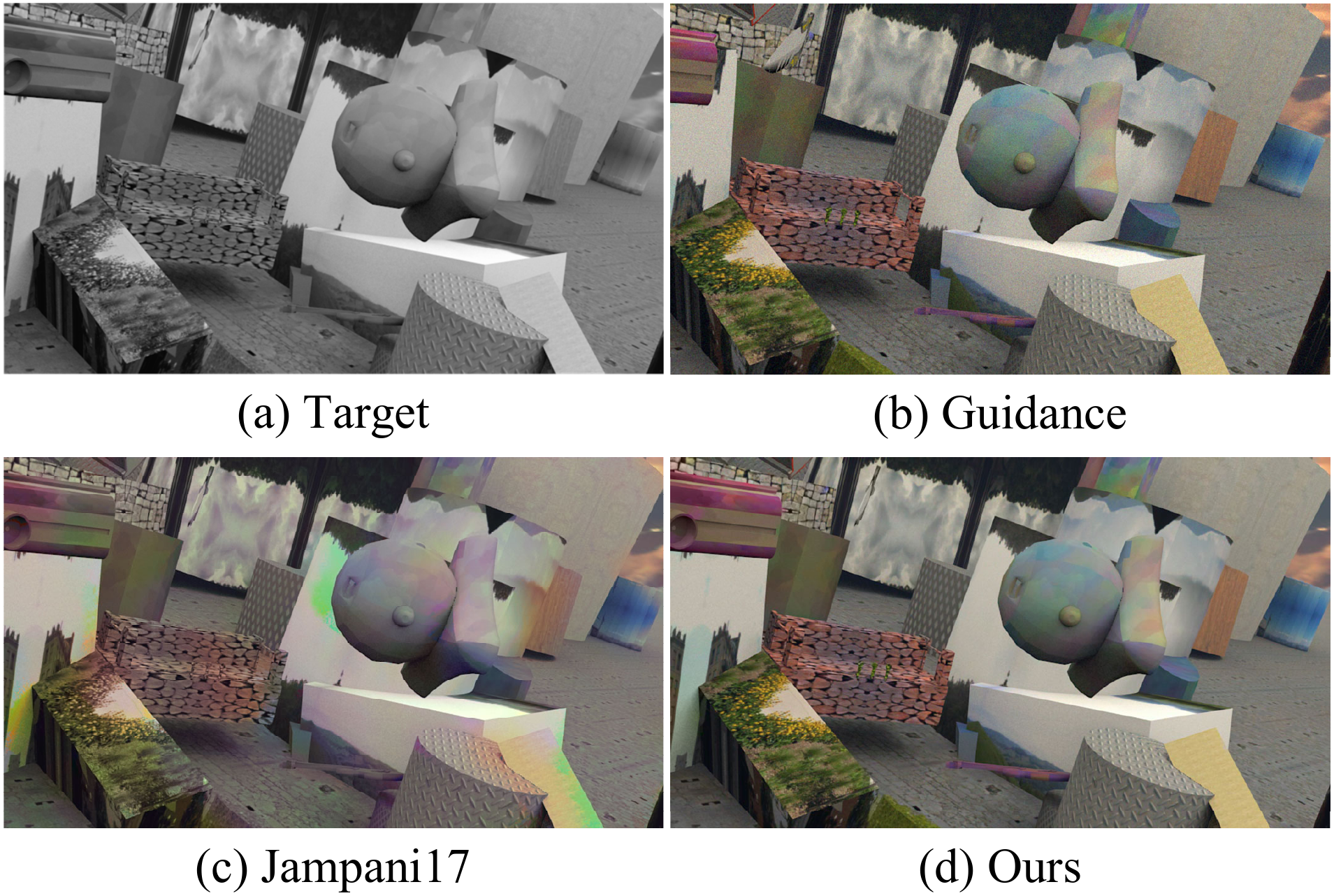}
\caption{Visual comparison of colorization results on the image pair from the SceneFlow test set, obtained by Jampani17~\cite{jampani2017video} and our algorithm, respectively.}
\label{fig:comp_sf}
\end{figure}

\subsection{Comparisons to Colorization Algorithms}
We quantitatively evaluate our algorithm on the Middlebury 2005, 2006 datasets~\cite{scharstein2007learning,hirschmuller2007evaluation} and the SceneFlow test set~\cite{mayer2016large}. We also compare it to the state-of-the-art example-based colorization algorithms including Jeon16~\cite{jeon2000stereo}, He18~\cite{he2018deep}, He19~\cite{he2019progressive} and Zhang19~\cite{zhang2019deep}, as well as the video color propagation algorithm Jampani17~\cite{jampani2017video}. To simulate image noise in real scenes, we add mixed Poisson-Gaussian noise~\cite{foi2008practical} to the guidance image with noise parameters $\alpha, \sigma^2 \in \{0, 0.03^2\}$, which indicate the noise variance of each pixel in the guidance image:

\begin{equation}\label{eq:noise_level}
\mathbf{\Gamma}(i,j,k) = \alpha \cdot \mathbf{C}(i,j,k) + \sigma^2.
\end{equation}

\begin{figure}[h]
\centering
\includegraphics[scale=0.205]{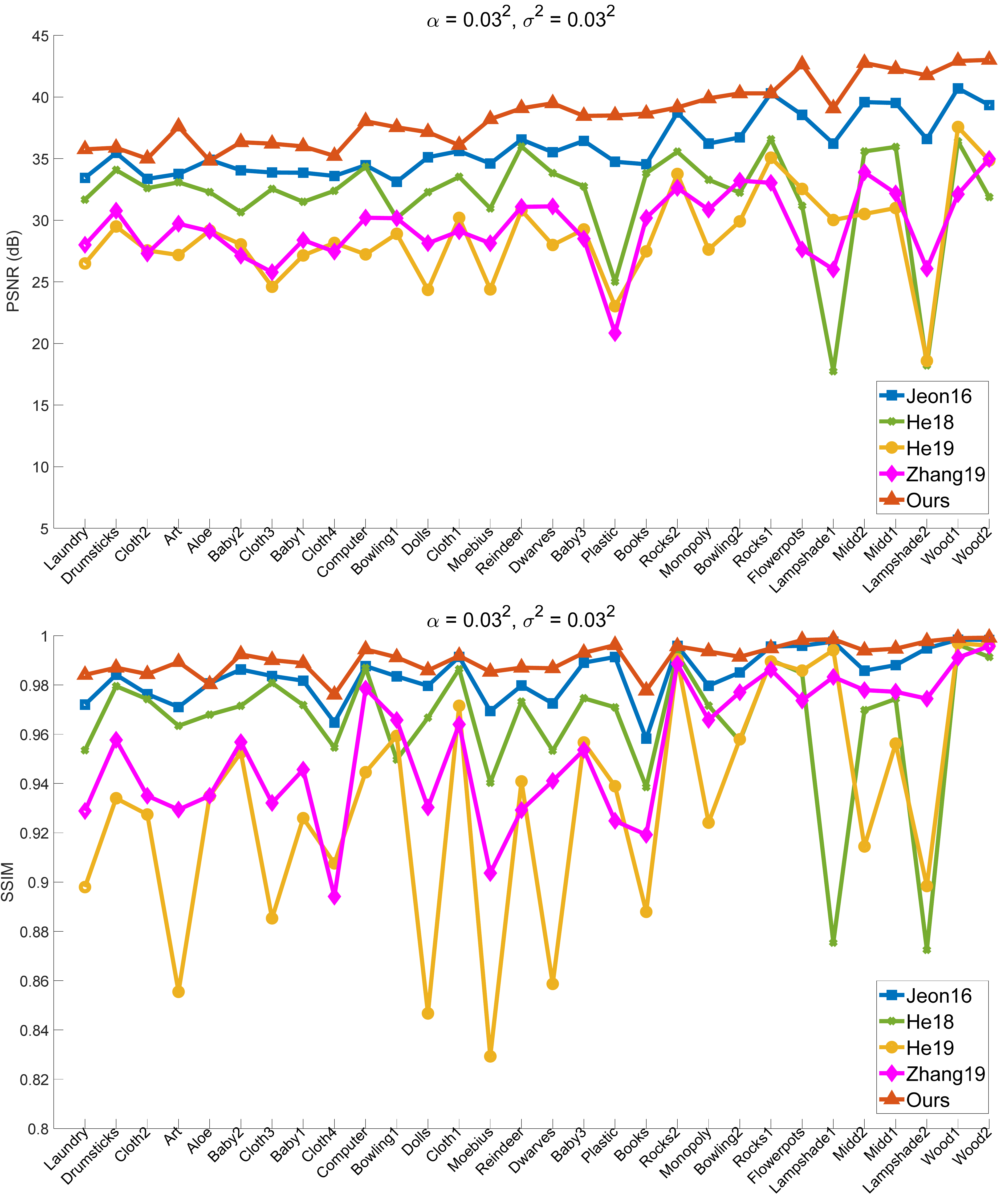}
\caption{The PSNR (dB) and SSIM values on each single image pair from the Middlebury datasets~\cite{scharstein2007learning,hirschmuller2007evaluation} obtained by Jeon16~\cite{jeon2000stereo}, He18~\cite{he2018deep}, He19~\cite{he2019progressive}, Zhang19~\cite{zhang2019deep} and our algorithm, respectively. The x-axis represents the image title, and the y-axis represents the corresponding PSNR and SSIM values.}
\label{fig:psnr_ssim}
\end{figure}

\begin{figure*}[t]
\centering
\includegraphics[scale=0.281]{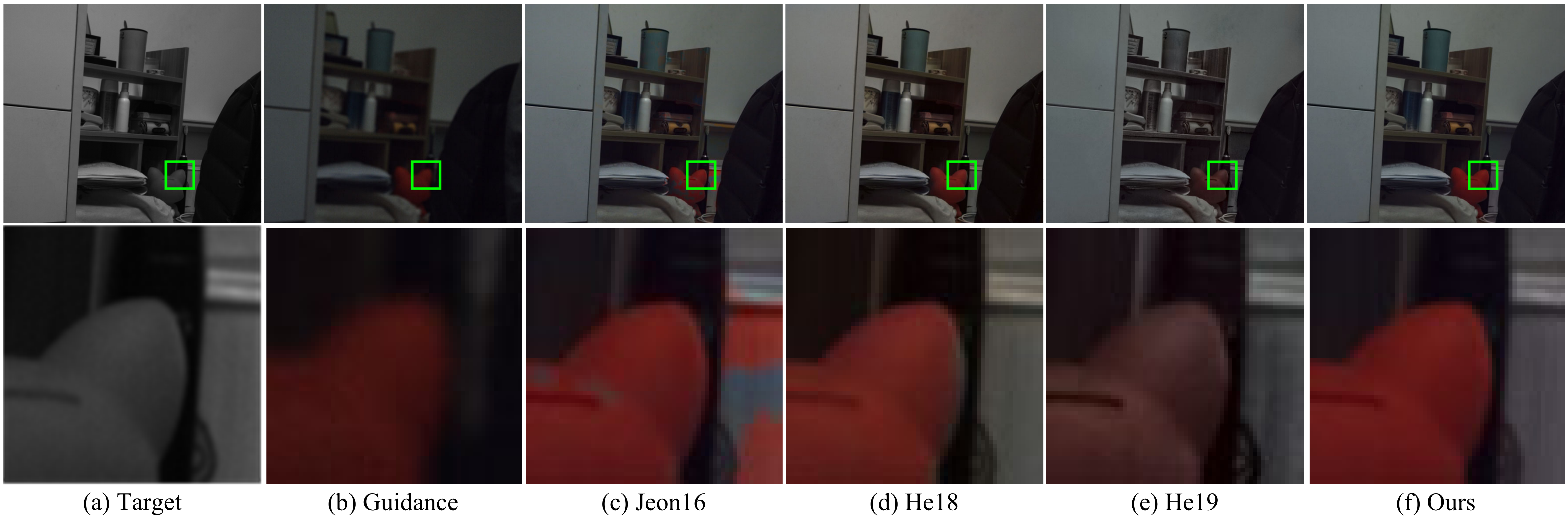}
\caption{Visual comparison of colorization results on the realistic image pairs, obtained by Jeon16~\cite{jeon2000stereo}, He18~\cite{he2018deep}, He19~\cite{he2019progressive}, and our algorithm, respectively.}
\label{fig:comp_real_1}
\end{figure*}

\begin{figure*}[h]
\centering
\includegraphics[scale=0.281]{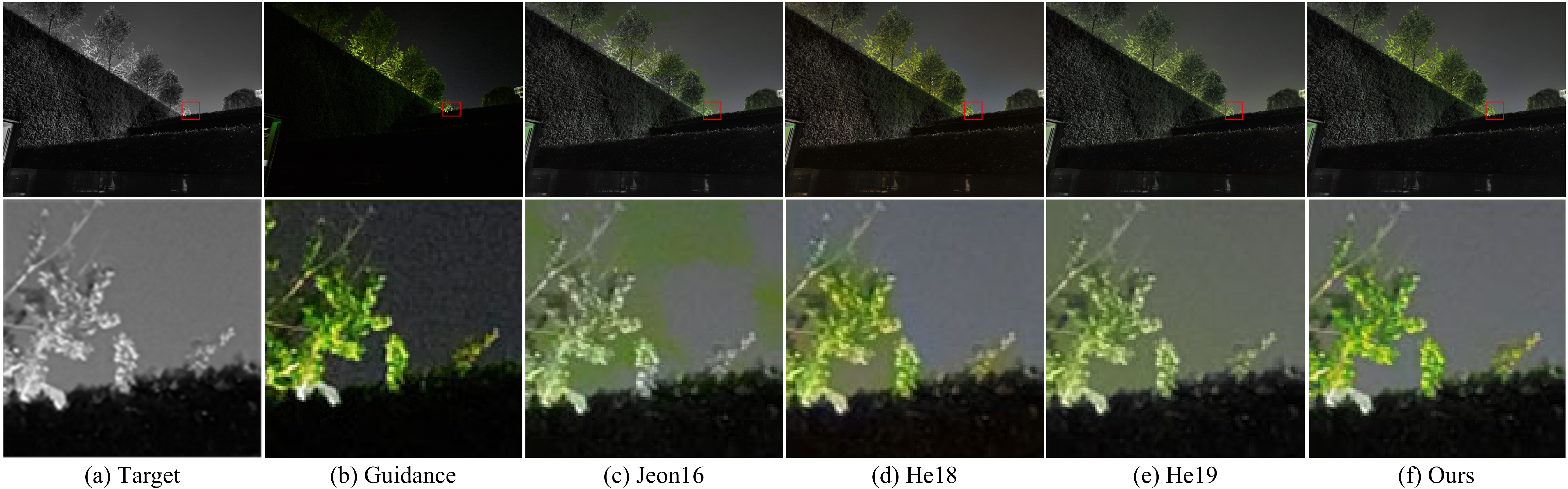}
\caption{Visual comparison of colorization results on the realistic image pairs, obtained by Jeon16~\cite{jeon2000stereo}, He18~\cite{he2018deep}, He19~\cite{he2019progressive}, and our algorithm, respectively.}
\label{fig:comp_real_2}
\end{figure*}

\begin{figure*}[h]
\centering
\includegraphics[scale=0.242]{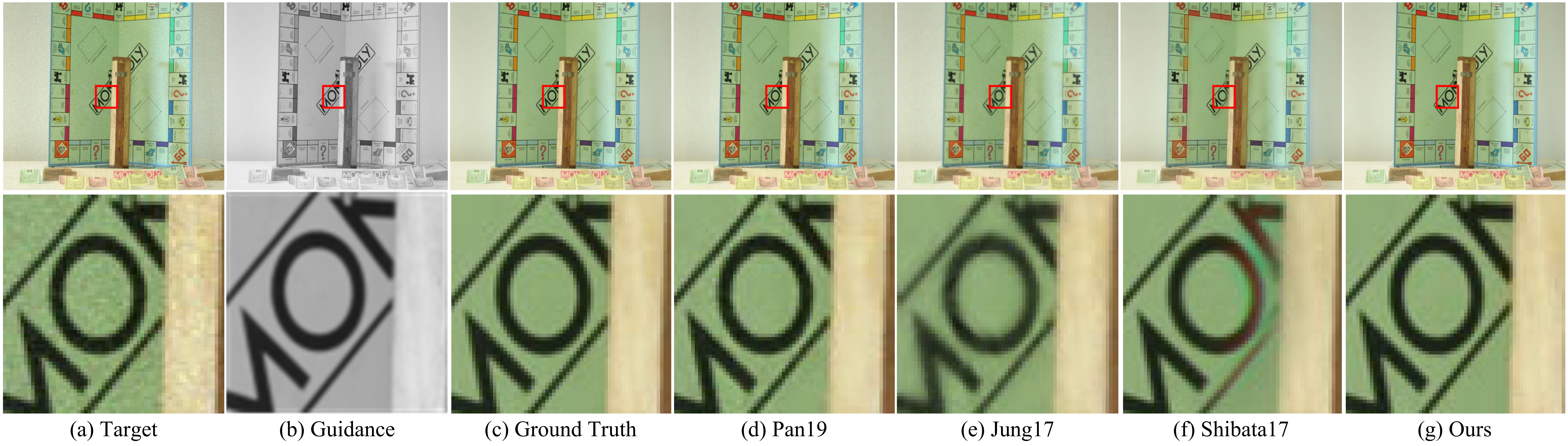}
\caption{Visual comparison of results obtained by our colorization algorithm and guided restoration algorithms including Pan19~\cite{pan2019spatially}, Jung17~\cite{jung2017enhancement}, and Shibata17~\cite{shibata2017misalignment}.}
\label{fig:comp_real_gr}
\end{figure*}

Results of He18~\cite{he2018deep}, He19~\cite{he2019progressive}, Zhang19~\cite{zhang2019deep} and Jampani17~\cite{jampani2017video} are obtained by the source codes and network models from the authors' websites, while results of Jeon16~\cite{jeon2000stereo} are obtained by our own implementation that follows the exact steps and parameter settings according to the paper. As feature matching can be directly performed between the target image and the lightness channel of the guidance image, decolorization of the guidance image in Jeon16~\cite{jeon2000stereo} are not included in the experiments. For fair comparison, deep learning methods are fine-tuned using 15000 image pairs from the SceneFlow train set. In addition, the lightness channel of each colorization result obtained by the compared algorithms is replaced by the target monochrome image as in our algorithm.

Table~\ref{tab:quantitative_result} and Table~\ref{tab:quantitative_result_sf} lists the average PSNR and SSIM obtained by different algorithms on the 30 mono-color image pairs from the Middlebury dataset. Visual results under noise parameters $\alpha=\sigma^2=0.03^2$ are shown in Fig.~\ref{fig:comp}. Jeon16~\cite{jeon2000stereo} achieves good PSNR and SSIM, and also shows plausible colorization results. It enhances the colorization method in~\cite{levin2004colorization} by introducing an additional weight term. The entire image is segmented into super-pixels and the weight term is computed by the median chrominance of each super-pixel. However, restricted by the accuracy of disparity estimation and super-pixel segmentation, color bleeding still exists. He18~\cite{he2018deep}, He19~\cite{he2019progressive} and Zhang19~\cite{zhang2019deep} obtain results with better visual quality in terms of less color bleeding. However, they do not achieve high PSNR and SSIM values. The main reason is that, they are designed to transfer colors between two dissimilar images. The use of perceptual loss causes the colorization results to be faithful to the overall style rather than the accurate colors of the guidance images. Jampani17~\cite{jampani2017video} propagates colors between two adjacent video frames where the motions are relatively small. Hence, it can be less effective when handling large disparity situations, as depicted in Fig. ~\ref{fig:comp_sf}. In comparison, our algorithm achieves the highest PSNR and SSIM values and also the best visual quality. As shown in Fig.~\ref{fig:psnr_ssim}, our algorithm basically out-performs other state-of-the-art algorithms on every image pair from the Middlebury datasets.

The codes of Jeon16~\cite{jeon2000stereo} and our algorithm are both implemented in MATLAB R2020b. The average size of the test images from the Middlebury datasets is $555\times 660$. Jeon16~\cite{jeon2000stereo} requires several minutes to colorize one monochrome image, which is also mentioned in the author's paper. In comparison, our algorithm only takes about 10.1 seconds per image (5.9 seconds for dense scribbling and 4.2 seconds for color propagation). Using C++ implementation and parallel computing can further reduce the runtime significantly. In comparison, the dense scribbling process without patch sampling takes about 184.1 seconds per image, which is computationally expensive.

In addition, we set up a dual-camera system to evaluate algorithms on real-world image pairs. Fig.~\ref{fig:comp_real_1} displays the colorization results of the scenes captured using mono and RGB cameras with identical settings. Fig.~\ref{fig:comp_real_2} shows the colorization results, where the source images are captured in the low-light environment using a smartphone equipped with the mono-color cameras. In comparison, colorization performance of each algorithm on the real-world data is similar to that on the synthetic image ones. Our algorithm can achieve colorization results with no color bleeding, even around tiny details, and restore colors faithfully according to the guidance image.

\subsection{Comparisons to Guided Restoration Algorithms}
As mentioned before, the other option to fuse a mono-color image pair is to denoise the color image under the guidance of the monochrome one. Hence, we also compare our algorithm to the state-of-the-art guided restoration methods. Pan19~\cite{pan2019spatially} performs self-guided denoising, while Jung17~\cite{jung2017enhancement} and Shibata17~\cite{shibata2017misalignment} are designed for restoration with unaligned image pairs. The visual comparison is displayed in Fig.~\ref{fig:comp_real_gr}. All the guided restoration methods can achieve plausible results. However, the self-guided Pan19~\cite{pan2019spatially} can lead to artifacts, which is also a common issue in single image denoising. Jung17~\cite{jung2017enhancement} causes the restoration results to be over-smooth though it contains a detail enhancement step. Shibata17~\cite{shibata2017misalignment} handles misalignment by shifting the guidance image, but it cannot resolve the misalignment problem caused by occlusion. It also requires a high amount of calculation if the displacement between two images is large. In comparison, the colorization option can produce images of higher visual quality in terms of better detail preservation and artifact control.

\subsection{Discussions}
\subsubsection{How Does Patch Sampling Influence The Accuracy?}
In this work, the size of each patch is set to $16\times 16$. We also set the sampling parameters $N=T=4$ so that $1/64$ of the patches are selected for dense scribbling, based on the two criteria $\Phi_\mathrm{valid\_match}(N,T)$ and $\Phi_\mathrm{confidence}(N,T)$. As mentioned, by setting $N=T=4$, we obtain $\Phi_\mathrm{valid\_match}(4,4)=0.7920$ and $\Phi_\mathrm{confidence}(4,4)=0.9639$. To show that $1/64$ of the patches are indeed sufficient for reliable color estimation, we perform guided colorization without patch sampling for comparison, which is equivalent to setting $N=256$. To colorize a similar number of pixels in dense scribbling, we set $T=214$. In this case, we obtain $\Phi_\mathrm{valid\_match}(256,214)=0.7987$ and $\Phi_\mathrm{confidence}(256,214)=0.9250$, which is close to the situation where $N=T=4$.

\begin{table}[tb!]\footnotesize
\renewcommand\arraystretch{1.2}
\renewcommand\tabcolsep{7pt}
\centering
\caption{The average PSNR (dB) and SSIM values on the image pairs from the Middlebury datasets~\cite{scharstein2007learning,hirschmuller2007evaluation} obtained by our colorization algorithm with and without patch sampling.}\label{tab:results_with_sampling}
\begin{tabular}{p{62 pt}|p{26 pt}<{\centering}|p{26 pt}<{\centering}|p{26 pt}<{\centering}|p{26 pt}<{\centering}}
  \hline\hline
  \multirow{3}{*}{} & \multicolumn{4}{c}{$\alpha=0$} \\\cline{2-5}
  &\multicolumn{2}{c|}{$\sigma^2=0$} & \multicolumn{2}{c}{$\sigma^2=0.03^2$} \\\cline{2-5}
  &PSNR&SSIM&PSNR&SSIM \\\hline
  w/o Sampling & 41.61 & 0.9955 & 40.37 & 0.9940 \\\cline{1-1}\cline{2-5}
  w/ Sampling & 41.50 & 0.9954 & 40.30 & 0.9938  \\\hline\hline
  \multirow{3}{*}{} & \multicolumn{4}{c}{$\alpha=0.03^2$} \\\cline{2-5}
  &\multicolumn{2}{c|}{$\sigma^2=0$} & \multicolumn{2}{c}{$\sigma^2=0.03^2$} \\\cline{2-5}
  &PSNR&SSIM&PSNR&SSIM \\\hline
  w/o Sampling & 39.03 & 0.9914 & 38.93 & 0.9913 \\\hline
  w/ Sampling & 38.81 & 0.9909 & 38.56 & 0.9904  \\\hline\hline
\end{tabular}
\end{table}

\textcolor{red}{
\begin{table}[tb!]\footnotesize
\renewcommand\arraystretch{1.2}
\renewcommand\tabcolsep{7pt}
\centering
\caption{The average PSNR (dB) and SSIM values on the image pairs from the Middlebury datasets~\cite{scharstein2007learning,hirschmuller2007evaluation} obtained by our colorization algorithm with and without denoising on the guidance image.}\label{tab:results_with_denoising}
\begin{tabular}{p{62 pt}|p{26 pt}<{\centering}|p{26 pt}<{\centering}|p{26 pt}<{\centering}|p{26 pt}<{\centering}}
  \hline\hline
  \multirow{3}{*}{} & \multicolumn{4}{c}{$\alpha=0$} \\\cline{2-5}
  &\multicolumn{2}{c|}{$\sigma^2=0$} & \multicolumn{2}{c}{$\sigma^2=0.03^2$} \\\cline{2-5}
  &PSNR&SSIM&PSNR&SSIM \\\hline
  w/o Denoising & \multirow{2}{*}{\textbf{41.50}} & \multirow{2}{*}{\textbf{0.9954}} & 40.26 & 0.9937 \\\cline{1-1}\cline{4-5}
  w/ Denoising & & & \textbf{40.30} &\textbf{0.9938}  \\\hline\hline
  \multirow{3}{*}{} & \multicolumn{4}{c}{$\alpha=0.03^2$} \\\cline{2-5}
  &\multicolumn{2}{c|}{$\sigma^2=0$} & \multicolumn{2}{c}{$\sigma^2=0.03^2$} \\\cline{2-5}
  &PSNR&SSIM&PSNR&SSIM \\\hline
  w/o Denoising & 35.52 & 0.9782 & 35.94 & 0.9817 \\\hline
  w/ Denoising & \textbf{38.81} & \textbf{0.9909} & \textbf{38.56} & \textbf{0.9904}  \\\hline\hline
\end{tabular}
\end{table}
}

Table~\ref{tab:results_with_sampling} lists the average PSNR and SSIM~\cite{wang2004image} values on the image pairs from the Middlebury datasets obtained by our colorization algorithm with and without patch sampling respectively. Their results are similar, but patch sampling obtains them in much shorter time, as reported above. Here, we observe that the PSNR and SSIM values obtained without sampling is slightly higher than those with sampling, but the former confidence value is lower than the latter one. The reason is that, the confidence criterion refers to the probability that a pixel obtains the correct color estimate judged by the JND threshold~\cite{wu2016enhanced}, while PSNR and SSIM measures the similarity between the estimate and the ground truth. If dense scribbling is conducted without patch sampling, more candidates can be averaged to generate a more accurate color estimate, leading to higher PSNR and SSIM values. However, although the color estimate obtained without sampling is closer to ground truth than that obtained with sampling, they are both judged to be correct by JND. In other words, human eyes cannot perceive the difference between the two colorization results, but patch sampling significantly reduces the amount of computation.

\subsubsection{How Does The Pre-denoising Step Influence The Accuracy?}
We use the random redundant DCT (RRDCT) denoising algorithm introduced in~\cite{fujita2015randomized} to remove the mixed Poisson-Gaussian noise in the guidance image. One of the major reasons we choose RRDCT is that it can runs in almost real time. Here, we give a rough introduction of the denoising process. The noisy image is first divided into multiple overlapped patches. For each patch, hard-thresholding is performed on its DCT coefficients for noise removal, with a threshold related to the noise variance. Noise intensity is approximated to be patch-wisely consistent, computed by the average noise variance of each patch. Finally, the denoised patches are aggregated to construct the complete denoised image.

Table~\ref{tab:results_with_denoising} shows the average PSNR and SSIM values on the 30 image pairs from the Middlebury datasets~\cite{scharstein2007learning,hirschmuller2007evaluation}, obtained by our colorization algorithm with and without denoising on the guidance image, respectively. It's clear that the pre-denoising step helps further improve the accuracy, especially when the noise level is relatively high. The main reason is that noise will affect the accuracy of block matching and cannot be completely eliminated by the weighted average of the color candidates. Spatial denoising is one of the most direct and effective manners to remove the noise of the color candidates. 

\begin{figure}[t]
\centering
\includegraphics[scale=0.273]{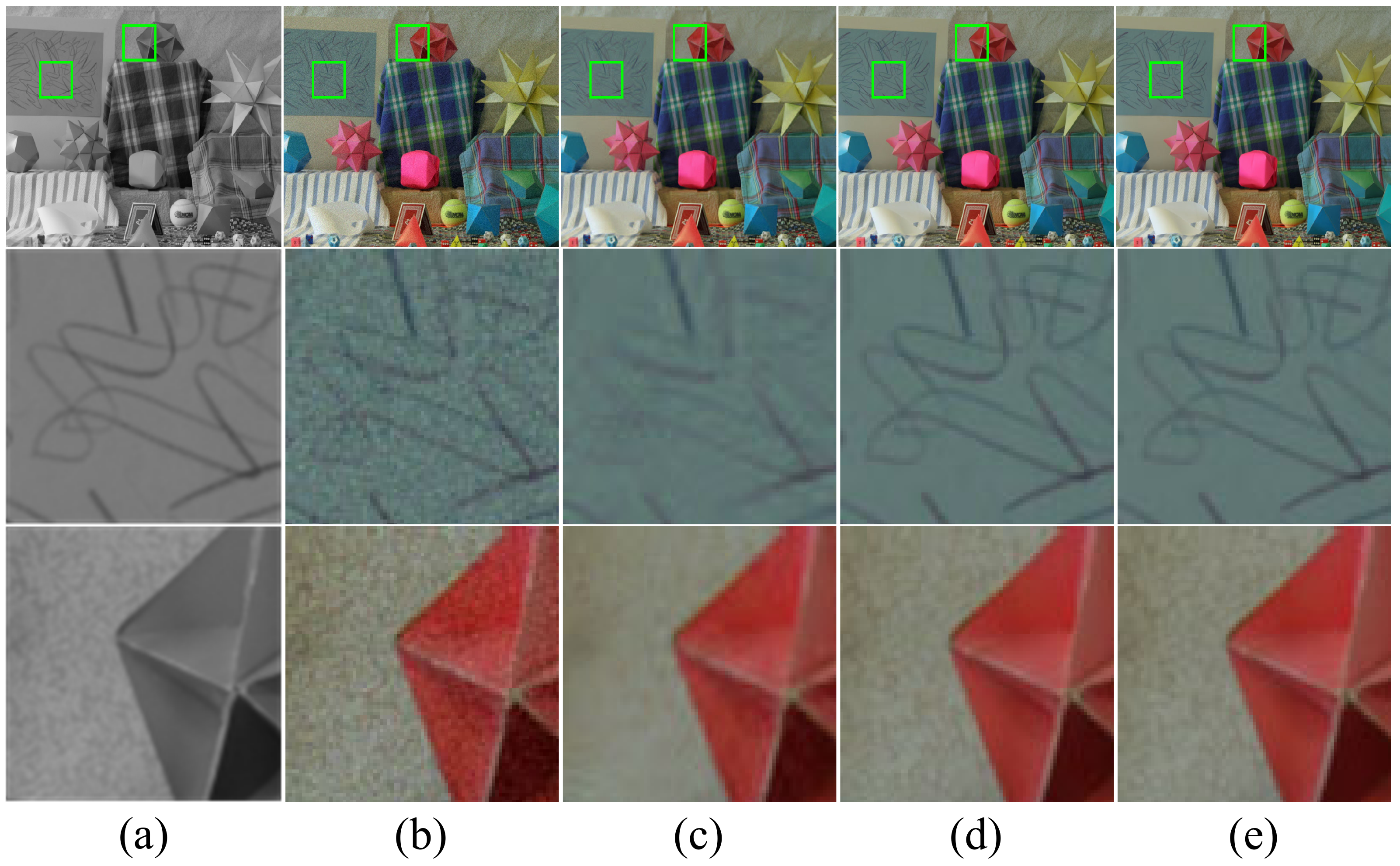}
\caption{Visual comparison between the denoised guidance images and the colorization results. (a) Monochrome target image. (b) Guidance image. (c) Denoised guidance image. (d) Colorization result. (e) Ground truth.}
\label{fig:comp_to_denoised}
\end{figure}

Fig.~\ref{fig:comp_to_denoised} (c) displays the denoised guidance images under the noise parameters $\alpha=\sigma^2=0.03^2$, while Fig.~\ref{fig:comp_to_denoised} (d) shows the corresponding colorization results. Visually, the denoised images suffer from artifacts and edge smoothing, but our colorization results are not effected by these problems thanks to the robustness of block matching. Hence, RRDCT is already sufficient for robust colorization, and it's not necessary to apply a more complicated denoising algorithm that increases the amount of calculation.

\section{Conclusions}
In this work, we propose a guided colorization to restore color images with better quality from the mono-color image pairs. The algorithm is composed of two stages: dense scribbling and color propagation. Dense scribbling applies block matching for robust color estimation. Two types of outliers including occluded and color-ambiguous pixels are detected and then removed from the initial dense scribbles. We also introduce a patch sampling strategy to accelerate the matching process. In the color propagation stage, we generate colors seed in the regions that have no color hints at all, and then propagate colors to the entire image. Experimental results show that, our algorithm is computationally efficient, and achieves good performance in solving the color bleeding problem.

\bibliographystyle{IEEEtran}
\bibliography{REF}

\end{document}